\documentclass[aps,prx,preprint,groupedaddress,longbibliography]{revtex4-2}
\usepackage[T1]{fontenc}

\usepackage{amsmath,amssymb,amsthm,mathtools,bm}
\usepackage{graphicx}
\usepackage{booktabs}
\usepackage{placeins}
\usepackage{algpseudocode}
\newcounter{algorithm}
\renewcommand{\thealgorithm}{\arabic{algorithm}}
\usepackage{microtype}
\usepackage{url}
\usepackage{hyperref}
\hypersetup{hidelinks}
\usepackage{tikz}
\usetikzlibrary{arrows.meta,positioning}
\usepackage{comment}
\usepackage{booktabs, siunitx, pgfplots}
\usetikzlibrary{calc,pgfplots.groupplots}
\pgfplotsset{compat=1.18}

\usepackage[dvipsnames]{xcolor}

\newtheorem{theorem}{Theorem}

\newtheorem{lemma}[theorem]{Lemma}
\newtheorem{corollary}[theorem]{Corollary}
\theoremstyle{definition}
\newtheorem{definition}[theorem]{Definition}
\newtheorem{assumption}[theorem]{Assumption}
\theoremstyle{remark}
\newtheorem{remark}[theorem]{Remark}

\newcommand{\cS}{\mathcal{S}}
\newcommand{\cX}{\mathcal{X}}

\newcommand{\cR}{\mathcal{R}}
\newcommand{\cP}{\mathcal{P}}

\newcommand{\ind}{\mathbf{1}}
\newcommand{\KL}{\mathrm{KL}}
\newcommand{\ESS}{\mathrm{ESS}}
\newcommand{\E}{\mathbb{E}}

\newcommand{\TV}{\mathrm{TV}}
\newcommand{\given}{\,\middle|\,}
\newcommand{\target}{\mu}
\newcommand{\prior}{Q}
\newcommand{\route}{R}
\newcommand{\paths}{\Gamma}

\begin{document}

\title{Sampling Decisions: Exact Path-Space Control for Physics-Informed Generative Sampling}

\author{Michael Chertkov}
\email{chertkov@arizona.edu}
\affiliation{Program in Applied Mathematics and Department of Mathematics, University of Arizona, Tucson, Arizona, USA}
\author{Hamidreza Behjoo}
\email{hamidreza.behjoo@gmail.com}
\affiliation{Program in Applied Mathematics and Department of Mathematics, University of Arizona, Tucson, Arizona, USA}
\author{Sungsoo Ahn}
\email{sungsoo.ahn@kaist.ac.kr}
\affiliation{Graduate School of AI, KAIST, Daejeon, Republic of Korea}

\begin{abstract}
Scientific generative models must turn tractable local decisions into globally correlated samples that respect physical constraints. We introduce \emph{Sampling Decisions}, a finite-horizon framework in which a structured object is assembled on a growing state graph and corrected globally by an exact path-space control law. For a prescribed Gibbs target, the corrected law is the unique relative-entropy projection of a sequential prior and is realized by a Doob $h$-transform with a linear backward desirability recursion. The same mathematical object can be read as a KL-optimal controller, a one-sided Schr\"odinger transport, and an ideal value or flow function for auto-regressive (with memory of the path) and GFlowNet-type generation. A route-resolved formulation yields a finite-particle algorithm, and we prove convergence of its transition kernels and terminal law as the path budget grows.

For binary graphical models, we prove a structural cancellation theorem: every fixed singleton-product prior disappears from the population correction, so even improved one-point marginals do not change the exact generative dynamics. The information that matters is conditional and prefix dependent. Statistical physics supplies such an inductive bias through a Local-Boltzmann prior, which absorbs interactions as spins are revealed; optimal path-space control then supplies the missing look-ahead field generated by the unrevealed subgraph. On exactly enumerable Ising grids, this physics-informed proposal increases effective sample size by factors ranging from about ten to nearly one thousand relative to product proposals and reaches the exact-target reference band at the tested budgets. Tests on a larger $10\times10$ grid (which is beyond exact enumeration ability) preserves the same hierarchy: Local-Boltzmann guidance avoids the severe weight collapse of the product proposals and approaches a long-run MCMC baseline on the reported diagnostics.

\end{abstract}
\maketitle

\section{Introduction}
\label{sec:intro}

Generative AI can be viewed as controlled probability transport: a tractable source of randomness is transformed into samples from a structured target law. In scientific applications, the target is often specified not by a normalized data density but by an energy or action,
\begin{equation}
\target(\sigma)=\frac{1}{Z}\,\bar\target(\sigma),
\qquad
\bar\target(\sigma)=\exp\left(-E(\sigma)\right),
\label{eq:target}
\end{equation}
where the partition function $Z$ and the required conditional free energies are generally unknown. Statistical physics supplies the energy landscape and its local interaction structure, but not an immediately tractable generative mechanism. Contemporary generative models realize probability transport through stochastic diffusion paths \citep{ho_denoising_2020,song_denoising_2022}, auto-regressive factorizations, or compositional flows on directed acyclic graphs such as Generative Flow Networks (GFlowNets) \citep{bengio_flow_2021,bengio_gflownet_2023}. The latter viewpoint is natural when a sample is assembled through a finite sequence of decisions: starting from an empty object, one reveals components until a terminal configuration $\sigma$ is obtained. In contrast, Markov chain Monte Carlo (MCMC) targets the same law as the stationary distribution of a process whose mixing time may be difficult to control.

The difficulty is that inexpensive local decisions generally produce the wrong global law. A heuristic policy may make plausible early choices yet accumulate terminal bias, whereas an exact chain-rule sampler requires conditional partition functions that can be as hard to compute as the original problem. \emph{Sampling Decisions} resolves this tension by separating a cheap generative prior from an exact correction in path space. The corrected process is the discrete finite-horizon analogue of a Doob transform and a linearly solvable stochastic controller \citep{e_todorov_linearly-solvable_2007,dvijotham_unifying_2012,kappen_path_2005}; equivalently, it is a one-sided entropy-regularized transport from the empty object to the prescribed terminal law. In generative-model language, the prior transition $q_t$ is a proposal policy, while the backward desirability $h_t$ is the ideal prefix value or flow that converts this proposal into an exact generator.

This formulation makes the interdisciplinary content precise. Statistical physics determines the terminal Gibbs weight and, through locality, suggests structured inductive biases. Stochastic optimal control and transport determine the least-informative global deformation of a reference process that enforces the target law. Generative AI supplies the compositional state graph and the approximation question: which policy, value function, or learned flow can represent the exact path-space correction efficiently? The present work answers this question analytically on finite graphs and uses particles rather than a neural network to expose the exact object that a scalable learned model would need to approximate.

The distinction between the population solution and its finite-particle implementation is essential. The exact transform can be written using the terminal marginal of the prior, but that marginal may remain combinatorially expensive even when individual prior transitions are simple. We therefore introduce a route-resolved formulation that specifies a target conditional law over construction routes and converts the correction into pathwise importance weights. This separates two questions that are often conflated:
\begin{enumerate}
    \item Which corrected path law is exact at the population level?
    \item Which prior proposal estimates that corrected law efficiently at finite path budget?
\end{enumerate}
The first is determined by the terminal and route constraints. The second is controlled by proposal overlap, prefix-conditioned weight degeneracy, and therefore by the information encoded in the generative policy.

Our main contributions are as follows.
\begin{enumerate}
\item \textbf{Exact path-space control for generative construction.}
We formulate sequential Gibbs sampling as a terminally constrained path-space KL projection on an auto-regressively growing state graph. The unique minimizer is a Doob $h$-transform whose desirability obeys a linear backward recursion. This identifies one object simultaneously as a linearly solvable controller, a one-sided Schr\"odinger transport, and an ideal value or flow function for a compositional generator.

\item \textbf{Route-resolved particle implementation and consistency.} By lifting the terminal target to an explicit law over complete construction paths, we obtain a direct weighted-count estimator of every corrected transition. We prove almost-sure convergence of the estimated kernels and convergence in total variation of the induced terminal law.

\item \textbf{A structural limit of singleton learning.} We define the Independent Order--Mark (IOM) class and prove that all fixed singleton mark probabilities cancel exactly from the canonical correction. Uniform, mean-field (MF), and belief-propagation (BP) proposals therefore produce the same population process whenever their ordering policy agrees. Accurate one-point prediction alone is not an expressive inductive bias for the corrected generator.

\item \textbf{Physics-informed Local-Boltzmann guidance.} We construct a prefix-dependent Ising prior from the heat-bath conditional on already revealed neighbors. Along every ordering, its numerators reproduce the complete Gibbs energy. Under a uniform route target, the residual path weight is only a product of local normalizers, and the exact corrected spin field separates into a revealed local field and a look-ahead field generated by the unrevealed subgraph.

\item \textbf{Mechanism-resolving numerical validation.} On three exactly enumerable Ising grids, we show that singleton-marginal accuracy is a poor predictor of finite-budget performance. The observed hierarchy is instead explained by effective sample size, with Local-Boltzmann guidance providing roughly one to nearly three orders of magnitude larger ESS than product proposals. A $10\times10$ experiment beyond exact enumeration preserves this hierarchy: LBP retains $5.1\%$ of the nominal path budget at $K=5\times10^4$, while every tested product proposal falls below $0.1\%$.

\end{enumerate}

The paper is organized as follows. Section~\ref{sec:setup} defines the growing-state generator and develops its connections to control, transport, diffusion models, GFlowNets, and particle methods. Section~\ref{sec:canonical} derives the canonical KL-optimal correction. Section~\ref{sec:particle} introduces route-resolved Sampling Decisions and proves finite-particle consistency. Section~\ref{sec:iom} establishes product-prior cancellation. Section~\ref{sec:lbp} develops physics-informed Local-Boltzmann guidance. Section~\ref{sec:experiments} reports the controlled Ising validation, and Sections~\ref{sec:limitations}--\ref{sec:conclusion} discuss scope and implications for learned scientific generators.

\section{Sequential sampling on a growing state graph}
\label{sec:setup}

This section formalizes the growing-state generator and then interprets the same path-space object through the languages of statistical physics, optimal control and transport, diffusion-based generation, GFlowNets, and particle inference.

\subsection{Prior paths and the terminal Gibbs law}\label{sec:prior-paths}

We start from the empty object, $S_0=s_0=\varnothing$, and construct one additional component at each generation step $t\in\{0,\ldots,T-1\}$. Let $\cS_t$ be the finite set of partial objects containing exactly $t$ revealed components. Thus, $S_t\in\cS_t$ records the complete history relevant for the next decision, while the state spaces themselves grow from the singleton $\cS_0=\{\varnothing\}$ to the terminal configuration space $\cS_T$. This is an auto-regressive construction. It is also Markov when formulated on the growing state graph, because the current partial object $S_t$ contains all previously revealed components and the next-step distribution depends on the past only through $S_t$.

\begin{figure}[t]
\centering
\includegraphics[width=0.52\textwidth]{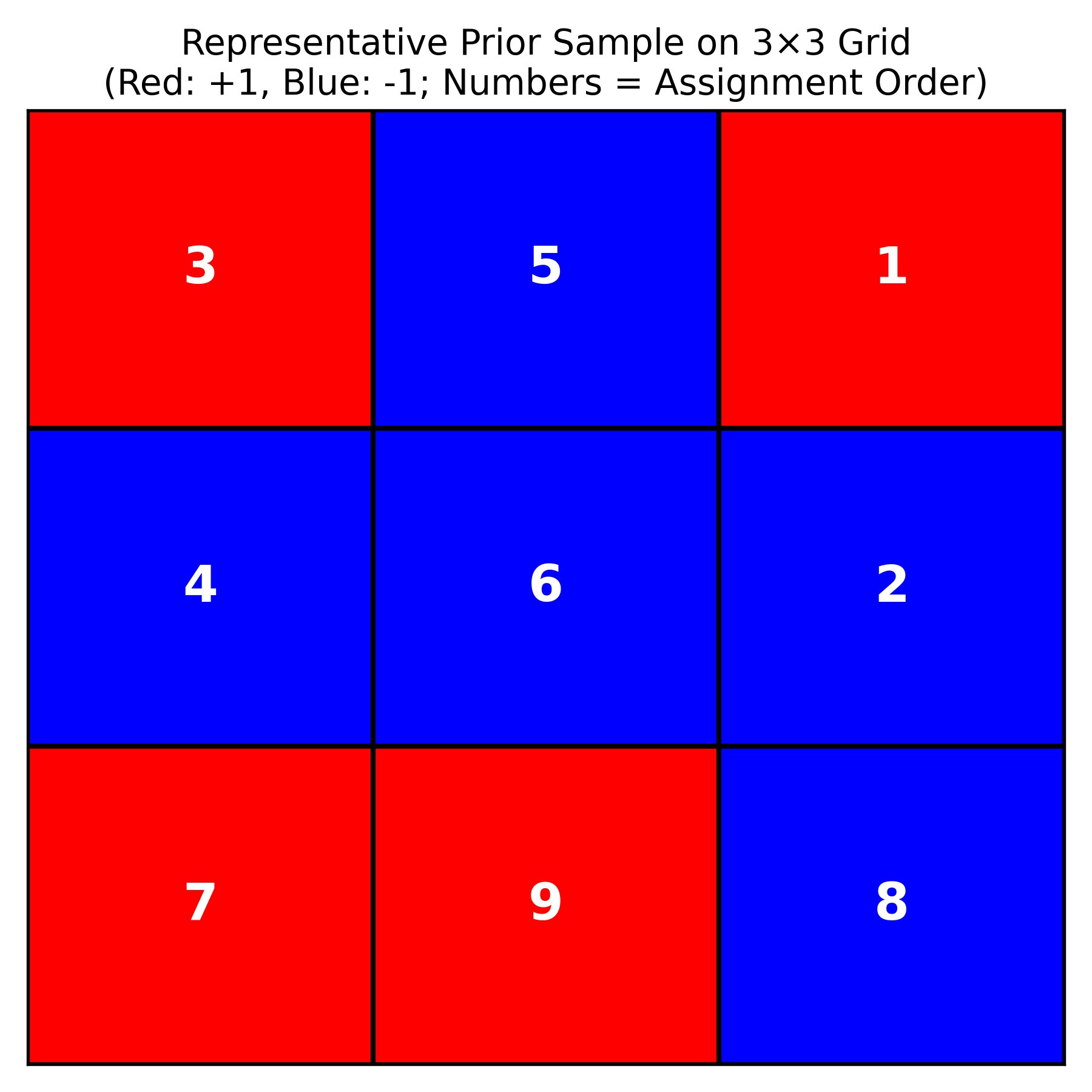}
\caption{A terminal Ising configuration together with one of its possible construction routes. Colors encode the terminal spins (red: $+1$, blue: $-1$), while the numbers indicate the order in which the sites are revealed. Multiple routes can lead to the same terminal configuration; distinguishing terminal states from construction routes is central to the path-space formulation of Sampling Decisions.}
\label{fig:construction-route}
\end{figure}

A prior construction process is specified by transition probabilities
\begin{equation}
q_t(s'\mid s),
\qquad s\in\cS_t,\quad s'\in\cS_{t+1},
\label{eq:priortransition}
\end{equation}
supported on the edges of the resulting layered directed acyclic graph. A complete construction route is
\[
\gamma=(s_0,s_1,\ldots,s_T)\in\paths,
\]
and its prior probability is
\begin{equation}
\prior(\gamma)=\prod_{t=0}^{T-1}q_t(s_{t+1}\mid s_t).
\label{eq:pathprior}
\end{equation}
Fig.~\ref{fig:construction-route} illustrates this distinction for an Ising configuration. The colors specify the completed spin configuration, whereas the numbers specify one particular order in which its components are revealed. Different reveal orders therefore define different paths through the growing state graph even when they produce the same terminal configuration.

The terminal state $S_T=\sigma$ belongs to the configuration space $\cX\doteq\cS_T$.  We denote the prior terminal marginal by
\begin{equation}
\prior_T(\sigma)=\sum_{\gamma:\,s_T=\sigma}\prior(\gamma).
\label{eq:priorterminal}
\end{equation}
Our goal is to construct a Markov path law whose terminal marginal is the Gibbs distribution in Eq.~\eqref{eq:target}.

\begin{assumption}[Support]
\label{ass:support}
Whenever $\bar\target(\sigma)>0$, the prior assigns positive terminal probability: $\prior_T(\sigma)>0$.
\end{assumption}

The fixed initial state and finite horizon are convenient rather than essential.  They cover the sequential assignment problems studied below and avoid measure-theoretic distractions.

With the prior path law, terminal marginal, and support condition now fixed, we next relate this growing-state construction to neighboring control, transport, generative-flow, and particle frameworks.

\subsection{Relation to neighboring frameworks}\label{sec:neighboring-frameworks}

\paragraph{Linearly solvable control and Doob transforms.}
Entropy-regularized control penalizes the deformation of a passive, or reference, transition kernel through a stage-wise KL cost. Under the exponential, or desirability, transformation, the corresponding Bellman recursion becomes linear \citep{kappen_path_2005,e_todorov_linearly-solvable_2007,dvijotham_unifying_2012, kappen_optimal_2012,theodorou_relative_2012}. Sampling Decisions uses the same mechanism but replaces a conventional terminal reward by the requirement that the terminal marginal equal a prescribed target distribution. Equivalently, it selects the path measure closest in relative entropy to the prior among all measures satisfying that terminal constraint. The resulting process is a time-inhomogeneous finite-state Doob transform \citep{doob_conditional_1957,chetrite_variational_2015}: its backward desirability is the prior conditional expectation of the terminal density ratio and propagates global target information to every prefix. In generative-model language, this desirability is an exact critic or value function, and the Doob ratio is the corresponding policy correction. Unlike standard fixed-dimensional control, the state graph grows with the object: layer $t$ records $t$ revealed variables and all information needed for the next decision.

\paragraph{Schr\"odinger bridges and entropy-regularized optimal transport.}
At the level of path measures, Sampling Decisions is a one-sided Schr\"odinger-type bridge, or equivalently an entropy-regularized transport problem with prescribed reference dynamics \citep{dai_pra_stochastic_1991,chen_optimal_2021,de_bortoli_diffusion_2021}. Classical Schr\"odinger bridges constrain both endpoint marginals and minimize relative entropy with respect to a reference process. Here the initial law is concentrated on the empty object and only the terminal Gibbs marginal is imposed. The transport therefore occurs on path space, not through a static Wasserstein coupling on one fixed-dimensional space. The reference process encodes admissible construction routes and inexpensive prior knowledge; the terminal constraint determines the smallest global reweighting needed for exact generation.

\paragraph{Path Integral Diffusion.} Diffusion and score-based generative models transport a simple reference law to a complex data law along a stochastic path \citep{ho_denoising_2020,song_denoising_2022}. The same change-of-measure principle appears in diffusion Schr\"odinger bridges \citep{de_bortoli_diffusion_2021} and in Path Integral Diffusion, where backward Green functions deform a stochastic prior to match a prescribed terminal law \citep{behjoo_harmonic_2025}; related control formulations provide guarantees for latent-diffusion generators \citep{tzen_theoretical_2019}. Sampling Decisions is a discrete-state counterpart, but not merely a time discretization. Its state space grows as components are revealed, and multiple construction routes can coalesce into the same terminal object. The path-space formulation must therefore represent both the completed sample and the route by which it was generated.

\paragraph{GFlowNets and auto-regressive sampling.} GFlowNets construct objects on a layered directed acyclic graph and learn forward and, in many formulations, backward flows so that terminal objects are sampled proportionally to a reward \citep{bengio_flow_2021,bengio_gflownet_2023,madan_learning_2023,deleu_discrete_2024}. Sampling Decisions shares the compositional graph and route multiplicity, but starts from an explicit prior path measure and derives the population correction by relative-entropy projection. The present paper does not train a neural flow. Instead it identifies the exact backward value, flow, and look-ahead quantities that a learned GFlowNet or auto-regressive policy would have to approximate, and it isolates which forms of prior information can reduce the residual correction.

\paragraph{Sequential importance sampling and twisted particles.} The route-resolved implementation in Section~\ref{sec:particle} approximates the backward desirability function using conditional, self-normalized importance sampling on path space. It is therefore closely related to Feynman--Kac particle constructions, sequential Monte Carlo samplers \citep{delmoral_sequential_2006}, and twisted particle filters \citep{whiteley_twisted_2014}. The distinctive feature here is that the particles represent routes through a growing combinatorial state graph and that their weights implement a terminal-marginal correction derived from the exact path-space problem. We do not claim that this finite-path implementation supersedes resampling-based sequential Monte Carlo. Rather, it provides a transparent particle realization of the population Doob transform and a controlled setting in which to diagnose how the choice of prior affects finite-budget performance.

\paragraph{One path-space object, three scientific roles.}
The construction can therefore be read in three complementary ways. Statistical physics specifies the target energy and the locality structure available to the prior. Optimal control and transport select the KL-minimal global correction and propagate it backward through a desirability function. Generative AI interprets the prior as a policy and the desirability as the ideal learned value or flow. This three-way identification is operational: it separates what can be supplied analytically by physics from what must be inferred, estimated, or amortized by a generative model.

\section{Sampling Decisions as a KL-optimal path correction}
\label{sec:canonical}

Every sequential construction defines a probability law $P$ on complete paths $\gamma\in\paths$. We compare $P$ with the prior path law $\prior$ defined in Eq.~\eqref{eq:pathprior}; its terminal marginal $\prior_T$ is given by Eq.~\eqref{eq:priorterminal}. Their path-space Kullback--Leibler divergence is
\begin{equation}
\KL(P\|\prior)
\doteq
\sum_{\gamma\in\paths}
P(\gamma)\log\frac{P(\gamma)}{\prior(\gamma)},
\label{eq:pathKL}
\end{equation}
with $\KL(P\|\prior)=+\infty$ if $P$ assigns positive probability to a path excluded by the prior. Let $\cP_\target$ denote the set of all path laws whose terminal marginal equals the prescribed target $\target$, that is,
\[
\sum_{\gamma:\,s_T=\sigma}P(\gamma)=\target(\sigma)
\qquad \text{for every }\sigma\in\cX.
\]
Sampling Decisions selects from $\cP_\target$ the path law that changes the prior as little as possible in the KL sense of Eq.~\eqref{eq:pathKL}. The admissible path laws need not initially be assumed Markov. If, however, $P$ is Markov on the growing state graph, with transitions $p_t(\cdot\mid s)$, then the chain rule for KL divergence gives
\begin{equation}
\KL(P\|\prior)
=
\sum_{t=0}^{T-1}\E_{S_t\sim P_t}
\left[
\KL\!\left(p_t(\cdot\mid S_t)\,\|\,q_t(\cdot\mid S_t)\right)
\right].
\label{eq:pathklstage}
\end{equation}
Here ``auto-regressive'' refers to the sequential construction, while ``Markov'' means that the current partial state contains all information from earlier decisions needed to choose the next component. The theorem below shows that the unique KL-minimizing path law indeed has this Markov form.
\begin{theorem}[Canonical path-space correction]
\label{thm:canonical}
Under Assumption~\ref{ass:support}, the optimization problem
\begin{equation}
P^\star
\in
\arg\min_{P\in\cP_\target}\KL(P\|\prior)
\label{eq:canonicaloptimization}
\end{equation}
has the unique solution
\begin{equation}
P^\star(\gamma)
=
\target(\sigma)\,\prior(\gamma\mid S_T=\sigma)
=
\prior(\gamma)\frac{\target(\sigma)}{\prior_T(\sigma)},
\qquad \sigma=s_T.
\label{eq:canonicalpathlaw}
\end{equation}
Moreover, $P^\star$ is Markov. Define the unnormalized terminal ratio and the backward desirability
\begin{align}
w_T(\sigma)&\doteq \frac{\bar\target(\sigma)}{\prior_T(\sigma)},
\label{eq:terminalweight}\\
h_t(s)&\doteq
\E_{\prior}\!\left[w_T(S_T)\mid S_t=s\right].
\label{eq:desirability}
\end{align}
Then $h_T=w_T$, and $h_t$ obeys the linear backward recursion
\begin{equation}
h_t(s)
=
\sum_{s'\in\cS_{t+1}}
q_t(s'\mid s)h_{t+1}(s').
\label{eq:linearbackward}
\end{equation}
The optimal transition kernel is the time-inhomogeneous Doob $h$-transform of the prior kernel $q_t$ \citep{doob_conditional_1957,chetrite_variational_2015}:
\begin{equation}
p_t^\star(s'\mid s)
=
q_t(s'\mid s)\frac{h_{t+1}(s')}{h_t(s)}.
\label{eq:doobtransition}
\end{equation}
Its terminal marginal is exactly $\target$.
\end{theorem}
\begin{proof}
Disintegrate any feasible path law as $P(\gamma)=\target(\sigma)P(\gamma\mid\sigma)$ and likewise write $\prior(\gamma)=\prior_T(\sigma)\prior(\gamma\mid\sigma)$.  The chain rule for relative entropy gives
\begin{align}
\KL(P\|\prior)
&=
\KL(\target\|\prior_T)
+
\sum_{\sigma\in\cX}\target(\sigma)
\KL\!\left(P(\cdot\mid\sigma)\,\|\,\prior(\cdot\mid\sigma)\right).
\label{eq:KLdecomp}
\end{align}
The first term is fixed by the terminal constraint.  The second KL term is nonnegative and vanishes only when $P(\cdot\mid\sigma)=\prior(\cdot\mid\sigma)$ on the target support, proving Eq.~\eqref{eq:canonicalpathlaw} and uniqueness.

For the transition formula, condition Eq.~\eqref{eq:canonicalpathlaw} on $S_t=s$.  The Radon--Nikodym factor depends only on $S_T$, so
\[
P^\star(S_{t+1}=s'\mid S_t=s)
=q_t(s'\mid s)
\frac{\E_{\prior}[w_T(S_T)\mid S_{t+1}=s']}
     {\E_{\prior}[w_T(S_T)\mid S_t=s]},
\]
which is Eq.~\eqref{eq:doobtransition}.  Summing over $s'$ yields Eq.~\eqref{eq:linearbackward}.  Finally, summing Eq.~\eqref{eq:canonicalpathlaw} over paths ending at $\sigma$ gives $P_T^\star(\sigma)=\target(\sigma)$.
\end{proof}

The backward function also has a Green-kernel representation.  Let
\begin{equation}
G^\prior_{t,T}(\sigma\mid s)
\doteq 
\prior(S_T=\sigma\mid S_t=s).
\label{eq:green}
\end{equation}
Unlike the reversed notation sometimes used for response functions, $G^\prior_{t,T}(\cdot\mid s)$ is a proper probability distribution over terminal states.  Eq.~\eqref{eq:desirability} becomes
\begin{equation}
h_t(s)
=
\sum_{\sigma\in\cX}
G^\prior_{t,T}(\sigma\mid s)
\frac{e^{-E(\sigma)}}{\prior_T(\sigma)}.
\label{eq:greenconvolution}
\end{equation}
Thus the canonical construction corrects the prior by convolving its forward terminal response with the target-to-prior terminal density ratio. In statistical-mechanics language, the terminal Boltzmann weight is converted into a prefix-dependent free-energy correction. In control language, this is the desirability. In generative language, it is the exact look-ahead value that turns a locally tractable proposal into a globally correct sampler.

\begin{corollary}[Linearly solvable MDP form]
\label{cor:lsmdp}
Among Markov transition kernels whose induced terminal marginal equals $\target$, Eq.~\eqref{eq:doobtransition} uniquely minimizes the stagewise KL cost in Eq.~\eqref{eq:pathklstage}.  Equivalently, with value $V_t(s)=-\log h_t(s)$, the Bellman recursion is linearized by the exponential transformation $h_t=e^{-V_t}$.
\end{corollary}

\begin{remark}[What is exact, and what must be computed]
Theorem~\ref{thm:canonical} is a population statement.  It is exact once $\prior_T$ or the conditional expectation in Eq.~\eqref{eq:desirability} can be evaluated.  A transition kernel can be analytically specified while its terminal marginal remains combinatorially expensive.  Section~\ref{sec:particle} addresses this gap without conflating an analytically defined prior with an exactly evaluated backward transform.
\end{remark}

\section{Route-resolved Sampling Decisions and finite-particle consistency}
\label{sec:particle}

This section turns the exact population value function of Section~\ref{sec:canonical} into a route-resolved estimator built from a finite bank of prior trajectories. The construction is deliberately nonparametric: it separates the mathematical target of learning from any particular neural approximation. We first lift the terminal target to a law over complete routes and derive its exact prefix-conditioned transitions. We then replace the corresponding prior expectations by weighted empirical counts, prove finite-particle consistency, and introduce ESS diagnostics for the finite-budget behavior studied in Section~\ref{sec:experiments}. In larger systems, the same weighted trajectories could instead supervise an amortized value, flow, or look-ahead model.

\subsection{Lifting the target from terminal states to paths}

The canonical path law in Eq.~\eqref{eq:canonicalpathlaw} retains the prior conditional distribution over routes given a terminal state.  This is KL optimal, but it requires the prior terminal marginal $\prior_T$.  A useful alternative is to specify the desired route conditional explicitly.

For each terminal configuration $\sigma$, let $\cR(\sigma)$ be the set of complete paths ending at $\sigma$, and let
\begin{equation}
\route(\gamma\mid\sigma),
\qquad \gamma\in\cR(\sigma),
\label{eq:routelaw}
\end{equation}
be any normalized route law.  Define the route-resolved target path distribution
\begin{equation}
P^{\route}(\gamma)
=
\target(\sigma)\route(\gamma\mid\sigma).
\label{eq:routetarget}
\end{equation}
Its terminal marginal is $\target$ by construction.  The canonical correction is recovered by choosing $\route(\gamma\mid\sigma)=\prior(\gamma\mid\sigma)$.

Assume $\prior(\gamma)>0$ whenever $P^{\route}(\gamma)>0$.  Up to the unknown global factor $Z^{-1}$, the pathwise importance ratio is
\begin{equation}
W_{\route}(\gamma)
\doteq 
\frac{\bar\target(\sigma)\route(\gamma\mid\sigma)}{\prior(\gamma)}.
\label{eq:pathweight}
\end{equation}
For any state $s\in\cS_t$ and child $s'\in\cS_{t+1}$, the exact route-resolved transition is
\begin{equation}
p_t^{\route}(s'\mid s)
=
\frac{
\E_{\prior}\!\left[W_{\route}(\Gamma)
\ind\{S_t=s,S_{t+1}=s'\}\right]
}{
\E_{\prior}\!\left[W_{\route}(\Gamma)
\ind\{S_t=s\}\right]
}.
\label{eq:populationweightedtransition}
\end{equation}
This is simply the conditional probability under $P^{\route}$, expressed as an expectation under the prior.

\begin{figure}[t]
\centering
\begin{tikzpicture}[
    node distance=7mm and 7mm,
    box/.style={draw, rounded corners, align=center, minimum width=0.285\textwidth, minimum height=10mm, inner sep=4pt, fill=gray!6},
    arrow/.style={-{Latex[length=2mm]}, thick}
]
\node[box] (priorbox) {Analytic sequential prior\\$\prior(\gamma)=\prod_t q_t(s_{t+1}\mid s_t)$};
\node[box, right=of priorbox] (bank) {$K$ prior construction paths\\$\Gamma^{(1)},\ldots,\Gamma^{(K)}$};
\node[box, right=of bank] (weights) {Route-resolved path weights\\$W_k\propto \bar\target(\sigma_k)\route(\Gamma^{(k)}\mid\sigma_k)/\prior(\Gamma^{(k)})$};
\node[box, below=of weights] (counts) {Prefix-conditioned\\weighted parent--child counts};
\node[box, left=of counts] (kernel) {Estimated corrected kernel\\$\widehat p_{t,K}^{\route}(s'\mid s)$};
\node[box, left=of kernel] (output) {Posterior construction paths\\terminal law $\to\target$ as $K\to\infty$};
\draw[arrow] (priorbox) -- (bank);
\draw[arrow] (bank) -- (weights);
\draw[arrow] (weights) -- (counts);
\draw[arrow] (counts) -- (kernel);
\draw[arrow] (kernel) -- (output);
\end{tikzpicture}
\caption{Route-resolved finite-particle Sampling Decisions. An analytically defined prior generates construction paths; route-resolved importance weights convert the path bank into prefix-conditioned weighted counts; these counts define an estimated corrected kernel whose terminal law converges to the target as the path budget grows.}
\label{fig:particle-workflow}
\end{figure}

Eq.~\eqref{eq:populationweightedtransition} is still a population identity. We now approximate its prior expectations using a finite bank of sampled construction paths.

\subsection{Finite path bank}

Draw $K$ independent prior paths $\Gamma^{(1)},\ldots,\Gamma^{(K)}\sim\prior$ and compute $W_k=W_{\route}(\Gamma^{(k)})$.  The direct particle estimator is
\begin{equation}
\widehat p_{t,K}^{\route}(s'\mid s)
=
\frac{
\sum_{k=1}^K W_k
\ind\{S_t^{(k)}=s,S_{t+1}^{(k)}=s'\}
}{
\sum_{k=1}^K W_k
\ind\{S_t^{(k)}=s\}
},
\label{eq:empiricaltransition}
\end{equation}

The estimator is defined whenever the denominator is positive. If the denominator vanishes, one may use any full-support fallback kernel; this finite-$K$ convention does not affect the asymptotic result below. In the numerical experiments of Section~\ref{sec:experiments}, the fallback was never invoked because every generated partial configuration retained at least one compatible continuation in the sampled path bank. This is an empirical observation rather than a guarantee, so the fallback remains necessary for a general finite path bank.

\begin{figure}[t]
\centering
\begin{minipage}{0.96\textwidth}
\refstepcounter{algorithm}
\noindent\textbf{Algorithm~\thealgorithm. Route-resolved finite-particle Sampling Decisions}\label{alg:particleDF}
\begin{algorithmic}[1]
\Require prior transitions $\{q_t\}_{t=0}^{T-1}$, unnormalized target $\bar\target$, route law $\route(\gamma\mid\sigma)$, path budget $K$
\State Sample $K$ independent prior paths $\Gamma^{(k)}=(S_0^{(k)},\ldots,S_T^{(k)})\sim\prior$
\State Compute $W_k=\bar\target(S_T^{(k)})\route(\Gamma^{(k)}\mid S_T^{(k)})/\prior(\Gamma^{(k)})$
\State Aggregate weighted parent--child counts for every $t$ and observed state
\State Form $\widehat p_{t,K}^{\route}$ using Eq.~\eqref{eq:empiricaltransition}; use a full-support fallback only when the denominator is zero
\State Generate posterior paths from $\{\widehat p_{t,K}^{\route}\}_{t=0}^{T-1}$
\end{algorithmic}
\end{minipage}
\end{figure}

The weighted-count form removes an ambiguity that arises if one first estimates a prior transition kernel and then inserts that estimate into a population Green-function formula.  Here the analytic prior remains the proposal, while the Monte Carlo approximation applies directly to the target conditional expectations.

\begin{theorem}[Finite-particle consistency]
\label{thm:consistency}
Suppose all state spaces are finite and $\prior(\gamma)>0$ on the support of $P^{\route}$.  Then, for every $t$, every state $s$ with $P^{\route}(S_t=s)>0$, and every child $s'$, 
\begin{equation}
\widehat p_{t,K}^{\route}(s'\mid s)
\longrightarrow
p_t^{\route}(s'\mid s)
\qquad \text{almost surely as }K\to\infty.
\label{eq:transitionconsistency}
\end{equation}
Because the collection of relevant states is finite, the convergence is simultaneous over all such transitions.  Let $\widehat P_{K,T}^{\route}$ be the terminal law generated by the estimated kernels.  Then
\begin{equation}
\left\|\widehat P_{K,T}^{\route}-\target\right\|_{\TV}
\longrightarrow 0
\qquad \text{almost surely}.
\label{eq:terminalconsistency}
\end{equation}
\end{theorem}

\begin{proof}
For fixed $(t,s,s')$, the numerator and denominator of Eq.~\eqref{eq:empiricaltransition}, divided by $K$, are empirical averages of nonnegative random variables with finite expectation.  The strong law of large numbers gives almost-sure convergence to the corresponding expectations in Eq.~\eqref{eq:populationweightedtransition}.  The denominator limit equals
\[
\E_{\prior}[W_{\route}(\Gamma)\ind\{S_t=s\}]
=Z\,P^{\route}(S_t=s)>0.
\]
The ratio therefore converges almost surely.  Finiteness allows intersection of the probability-one events over all relevant transitions.  The terminal probability of any $\sigma$ is a finite sum of finite products of transition probabilities, hence is continuous in the transition arrays.  Pointwise convergence on the finite terminal space implies total-variation convergence.
\end{proof}

\begin{remark}[Dependence of generated posterior samples]
Conditional on the shared path bank, posterior paths generated from $\widehat p_{t,K}^{\route}$ are independent.  Unconditionally, they share the randomness of the estimated kernel and are not exact i.i.d. target samples at finite $K$.  Theorem~\ref{thm:consistency} states that their common conditional terminal law converges to the target.
\end{remark}

The finite path bank can therefore be viewed as an inference-time approximation to the exact critic $h_t$. It is not presented as the final scalable architecture; rather, it supplies an analytically defined training target and a diagnostic baseline for learned or resampling-based approximations.

\subsection{Weight degeneracy and effective sample size}

Theorem~\ref{thm:consistency} establishes asymptotic exactness but does not determine the path budget required in practice. To connect the particle construction to the finite-budget behavior reported in Section~\ref{sec:experiments}, we diagnose the concentration of weights both globally and within the prefix-conditioned subsets used in Eq.~\eqref{eq:empiricaltransition}. We report the standard global diagnostic
\begin{equation}
\ESS
=
\frac{\left(\sum_{k=1}^K W_k\right)^2}
     {\sum_{k=1}^K W_k^2},
\qquad 1\le \ESS\le K,
\label{eq:ess}
\end{equation}
and, conceptually, the more relevant conditioned quantity
\begin{equation}
\ESS_t(s)
=
\frac{\left(\sum_{k=1}^K W_k\ind\{S_t^{(k)}=s\}\right)^2}
     {\sum_{k=1}^K W_k^2\ind\{S_t^{(k)}=s\}}.
\label{eq:conditionaless}
\end{equation}

\section{Independent Order--Mark priors and exact cancellation}
\label{sec:iom}

We now specialize to binary graphical models.  Let $V=\{1,\ldots,N\}$ be the variable set.  A partial state is
\[
s_t=(V_t,x_{V_t}),
\qquad |V_t|=t,
\]
where $V_t$ is the set of revealed variables and $x_{V_t}\in\{-1,+1\}^{V_t}$.  A child $s_t^{i,a}$ is obtained by selecting $i\notin V_t$ and assigning $x_i=a\in\{-1,+1\}$.

\begin{definition}[Independent Order--Mark prior]
\label{def:iom}
An IOM prior has transitions
\begin{equation}
q_t^{\mathrm{IOM}}(s_t^{i,a}\mid s_t)
=
\nu_i(V_t)\,q_i(a),
\label{eq:iomtransition}
\end{equation}
where $\nu_i(V_t)\ge0$, $\sum_{i\notin V_t}\nu_i(V_t)=1$, depends only on the revealed set and not on its values, while $q_i$ is a fixed full support singleton distribution.
\end{definition}

Uniform random ordering corresponds to $\nu_i(V_t)=1/(N-t)$.  A Plackett--Luce rule with positive scores $r_i$ corresponds to $\nu_i(V_t)=r_i/\sum_{j\notin V_t}r_j$ \citep{plackett_analysis_1975}.  The terminal prior factorizes:
\begin{equation}
\prior_T^{\mathrm{IOM}}(\sigma)=\prod_{i=1}^N q_i(\sigma_i).
\label{eq:iomterminal}
\end{equation}
Indeed, the order probabilities sum to one and the product of marks is order independent.

\begin{theorem}[Product-mark cancellation]
\label{thm:iom}
For an IOM prior, the canonical corrected process has transition
\begin{equation}
p_t^\star(i,a\mid s_t)
=
\nu_i(V_t)\,
\target\!\left(\sigma_i=a\given \sigma_{V_t}=x_{V_t}\right).
\label{eq:iomcancel}
\end{equation}
In particular, the corrected population dynamics is independent of every singleton mark $q_i$.  The prior ordering kernel $\nu$ survives unchanged.
\end{theorem}

\begin{proof}
For a complete assignment $\sigma$ and ordering $\pi$, the IOM path law factorizes as
\[
\prior(\pi,\sigma)
=
\prior_{\mathrm{ord}}(\pi)\prod_{i=1}^Nq_i(\sigma_i),
\]
where $\prior_{\mathrm{ord}}(\pi)=\prod_t\nu_{\pi(t+1)}(V_t^\pi)$.  Hence
$\prior(\pi\mid\sigma)=\prior_{\mathrm{ord}}(\pi)$ is independent of $\sigma$.  By Theorem~\ref{thm:canonical},
\[
P^\star(\pi,\sigma)=\target(\sigma)\prior_{\mathrm{ord}}(\pi).
\]
Conditioned on a partial assignment, the next node is selected according to $\nu_i(V_t)$ and its value follows the target conditional, proving Eq.~\eqref{eq:iomcancel}.
\end{proof}

The theorem is a structural statement about generative representation within the IOM class. Improving independently trained singleton logits cannot add expressive information to the population-corrected generator: mean-field, belief-propagation, and uniform marks all disappear when their order policies agree. Singleton accuracy is therefore not a sufficient objective for path-space generative design. A useful inductive bias must depend on the current prefix, alter the ordering policy, or both. The singleton proposals can nevertheless have very different finite-$K$ behavior because, under the canonical route law, the path weight becomes
\begin{equation}
W_{\mathrm{IOM}}(\pi,\sigma)
\propto
\frac{e^{-E(\sigma)}}{\prod_{i=1}^Nq_i(\sigma_i)}.
\label{eq:iomweight}
\end{equation}
A product proposal that fits singleton moments accurately may still under-cover the target's correlated modes.  Conversely, an overpolarized mean-field solution can assign extremely small mass to relevant configurations and create catastrophic weights.

\begin{remark}[Ordering remains a genuine design variable]
Theorem~\ref{thm:iom} removes only the fixed marks.  Different order kernels produce different intermediate path laws even though every exact process has the same terminal target.  In particular, a fixed order gives the usual target chain rule for that order, whereas random order gives a mixture of chain-rule factorizations.
\end{remark}

\section{Local-Boltzmann guidance for Ising models}
\label{sec:lbp}

\subsection{A prefix-dependent prior}

The Local-Boltzmann prior is a physics-informed generative policy. It uses the Hamiltonian's local decomposition to incorporate each interaction when it first becomes observable along the construction path, while leaving the influence of unrevealed variables to the global path-space correction.

Consider an Ising model on an undirected graph $G=(V,\mathcal E)$:
\begin{equation}
E(\sigma)
=-\sum_{(i,j)\in\mathcal E}J_{ij}\sigma_i\sigma_j
 -\sum_{i\in V}h_i\sigma_i.
\label{eq:isingenergy}
\end{equation}
For a partial assignment $s_t=(V_t,x_{V_t})$, define the field contributed by the external field and already revealed neighbors,
\begin{equation}
h_i^{\mathrm{loc}}(s_t)
=
h_i+
\sum_{j\in V_t:\,(i,j)\in\mathcal E}J_{ij}x_j.
\label{eq:localfield}
\end{equation}
The Local-Boltzmann prior (LBP) selects an unassigned node uniformly and samples its spin from the heat-bath conditional based only on revealed neighbors:
\begin{equation}
q_t^{\mathrm{LBP}}(i,a\mid s_t)
=
\frac{1}{N-t}
\frac{\exp[a h_i^{\mathrm{loc}}(s_t)]}
     {2\cosh h_i^{\mathrm{loc}}(s_t)}.
\label{eq:lbptransition}
\end{equation}
This is a one-pass sequential analogue of pseudo-likelihood sampling.  It is not a Gibbs sampler: each site is visited once, and unrevealed neighbors are omitted.  Consequently, its terminal law is generally biased.

\begin{lemma}[Order-wise energy absorption]
\label{lem:energyabsorption}
For every complete configuration $\sigma$ and every ordering $\pi$ of the nodes,
\begin{equation}
\prod_{t=0}^{N-1}
\exp\!\left(
\sigma_{\pi(t+1)}
 h_{\pi(t+1)}^{\mathrm{loc}}(s_t^\pi)
\right)
=
\exp(-E(\sigma)).
\label{eq:energyabsorption}
\end{equation}
\end{lemma}

\begin{proof}
Every field term $h_i\sigma_i$ appears when node $i$ is assigned.  For each edge $(i,j)$, the interaction $J_{ij}\sigma_i\sigma_j$ appears exactly once, when the later endpoint in the ordering is assigned and the earlier endpoint is already revealed.  The accumulated exponent is therefore $\sum_i h_i\sigma_i+\sum_{(i,j)}J_{ij}\sigma_i\sigma_j=-E(\sigma)$.
\end{proof}

Define the order-dependent product of local normalizers
\begin{equation}
Z_\pi^{\mathrm{loc}}(\sigma)
\doteq 
\prod_{t=0}^{N-1}
2\cosh\!\left(h_{\pi(t+1)}^{\mathrm{loc}}(s_t^\pi)\right).
\label{eq:zpiloc}
\end{equation}
Combining Eqs.~\eqref{eq:lbptransition} and \eqref{eq:energyabsorption} gives the complete LBP path probability
\begin{equation}
\prior_{\mathrm{LBP}}(\pi,\sigma)
=
\frac{1}{N!}\,
\frac{e^{-E(\sigma)}}{Z_\pi^{\mathrm{loc}}(\sigma)}.
\label{eq:lbppathlaw}
\end{equation}
The full energy is already present in the proposal; only the sequence of local normalizers distorts the path law.

\subsection{Uniform-route correction and the look-ahead field}

For computation we choose the target route law to be uniform and independent of the terminal configuration,
\begin{equation}
\route_{\mathrm{u}}(\pi\mid\sigma)=\frac{1}{N!}.
\label{eq:uniformroute}
\end{equation}
The desired path distribution is therefore $P^{\mathrm{u}}(\pi,\sigma)=\target(\sigma)/N!$.  From Eqs.~\eqref{eq:pathweight} and \eqref{eq:lbppathlaw}, the unnormalized LBP path weight simplifies exactly to
\begin{equation}
W_{\mathrm{LBP}}(\pi,\sigma)
=Z_\pi^{\mathrm{loc}}(\sigma).
\label{eq:lbpweight}
\end{equation}
Thus the bare Boltzmann factor cancels pathwise.  This does not imply dimension-free variance---a product of local normalizers can still fluctuate strongly---but it removes the full energy mismatch carried by a generic product proposal.

Under the uniform route target, node order is independent of $\sigma$.  The exact population transition is consequently
\begin{equation}
p_t^{\mathrm{u}}(i,a\mid s_t)
=
\frac{1}{N-t}
\target\!\left(\sigma_i=a\given \sigma_{V_t}=x_{V_t}\right).
\label{eq:uniformroutechainrule}
\end{equation}
This common exact limit is independent of the proposal used to estimate it.  The role of LBP is finite-particle preconditioning.

To expose the correction supplied by Sampling Decisions, define the target conditional magnetization
\begin{equation}
m_i(s_t)
=
\E_{\target}\!\left[
\sigma_i\given \sigma_{V_t}=x_{V_t}
\right]
\label{eq:conditionalmagnetization}
\end{equation}
and the look-ahead field
\begin{equation}
\widetilde h_i(s_t)
=
\operatorname{atanh}m_i(s_t)-h_i^{\mathrm{loc}}(s_t).
\label{eq:lookahead}
\end{equation}
Then Eq.~\eqref{eq:uniformroutechainrule} factorizes as
\begin{equation}
p_t^{\mathrm{u}}(i,a\mid s_t)
=
\frac{1}{N-t}
\frac{\exp\left(a(h_i^{\mathrm{loc}}(s_t)+\widetilde h_i(s_t))\right)}
     {2\cosh\left(h_i^{\mathrm{loc}}(s_t)+\widetilde h_i(s_t)\right)}.
\label{eq:correctedfield}
\end{equation}
The prior supplies the mechanistically available field of revealed neighbors, while the path-space correction supplies exactly the missing log-odds contribution of the unrevealed subgraph. The latter is a conditional free-energy difference in statistical physics, a value correction in stochastic control, and an ideal look-ahead signal for a learned generative policy. Equivalently,
\begin{equation}
\widetilde h_i(s_t)
=
\frac12\log
\frac{
\sum_{\sigma:\,\sigma\succ s_t,\,\sigma_i=+1}e^{-E(\sigma)}
}{
\sum_{\sigma:\,\sigma\succ s_t,\,\sigma_i=-1}e^{-E(\sigma)}
}
-h_i^{\mathrm{loc}}(s_t).
\label{eq:lookaheadpartition}
\end{equation}
The exact expression is generally intractable, but Algorithm~\ref{alg:particleDF} estimates the corresponding transitions from weighted LBP paths.

\subsection{Canonical versus route-resolved LBP}

The uniform-route correction used above should be distinguished from the canonical terminal-only KL projection.  Summing Eq.~\eqref{eq:lbppathlaw} over orderings gives
\begin{equation}
\prior_{\mathrm{LBP},T}(\sigma)
=
e^{-E(\sigma)}A_0(\sigma),
\qquad
A_0(\sigma)=\frac{1}{N!}\sum_{\pi}\frac{1}{Z_\pi^{\mathrm{loc}}(\sigma)}.
\label{eq:lbpterminal}
\end{equation}
The canonical terminal-only correction would use the terminal ratio $1/A_0(\sigma)$ and would retain the prior's configuration-dependent conditional law over orderings.  Computing $A_0$ requires an order average with up to $N!$ terms.  The uniform-route lift instead specifies the route law directly, gives the tractable path weight in Eq.~\eqref{eq:lbpweight}, and preserves uniform order in the exact corrected process.  Both constructions have the exact Gibbs terminal marginal; only the former is the KL-minimizer under a terminal-only constraint.  Appendix~\ref{app:canonicalLBP} gives the canonical LBP formulas.

\section{Controlled numerical validation}
\label{sec:experiments}

\subsection{Questions and experimental protocol}

The experiments are designed as a mechanism-resolving validation rather than a large-scale benchmark. Exact enumeration supplies a gold-standard target and allows us to distinguish a theorem about the population generator from the finite-particle efficiency of the proposal used to estimate it. We test the specific prediction that priors with the same exact corrected process can require sharply different path budgets when their residual importance weights have different variance.

We consider ferromagnetic nearest-neighbor Ising grids of size $3\times3$, $4\times4$, and $5\times5$ ($N=9,16,25$).  Couplings are sampled as $J_{ij}=0.3+U[0,0.2]$ and fields as $h_i\sim U[-0.1,0.1]$, with one fixed instance at each size.  Exact partition functions, singleton means, and pair correlations are obtained by enumeration of all $2^N$ configurations.  For each prior and path budget $K$, Algorithm~\ref{alg:particleDF} constructs an estimated uniform-route corrected kernel and generates $S=K$ posterior samples.  Three independent algorithmic seeds are used.

The proposals are:
\begin{itemize}
    \item \textbf{Uniform IOM:} $q_i(+1)=q_i(-1)=1/2$.
    \item \textbf{MF IOM:} $q_i$ is obtained from the naive mean-field fixed point.
    \item \textbf{BP IOM:} $q_i$ is obtained from belief-propagation singleton estimates.
    \item \textbf{LBP:} the prefix-dependent local conditional in Eq.~\eqref{eq:lbptransition}.
\end{itemize}
All use uniform node order in both the proposal and the route-resolved target.  The IOM path weights are given by Eq.~\eqref{eq:iomweight}; the LBP path weights are given by Eq.~\eqref{eq:lbpweight}.  Appendix~\ref{app:mf-bp} records the MF and BP equations.

For reference, the figures include (i) $5000$ i.i.d. samples drawn directly from the enumerated target and (ii) a single-spin Metropolis chain with $2000$ burn-in sweeps and thinning by $10$ sweeps.  The legend in the figures calls the direct target reference ``Exact-DF''; throughout the text we call it the \emph{exact target i.i.d. reference}.  MCMC is included only as context.  We do not claim a cost-normalized superiority result over MCMC.

\subsection{Diagnostics}

For posterior samples $\{\sigma^{(r)}\}_{r=1}^S$, define singleton and pair estimates
\begin{equation}
\widehat m_i=\frac1S\sum_{r=1}^S\sigma_i^{(r)},
\qquad
\widehat c_{ij}=\frac1S\sum_{r=1}^S\sigma_i^{(r)}\sigma_j^{(r)}.
\label{eq:moments}
\end{equation}
The reported relative discrepancies are
\begin{align}
\Delta_1
&=
\frac1N\sum_i
\frac{|\widehat m_i-m_i^{\mathrm{ref}}|}{|m_i^{\mathrm{ref}}|},
\label{eq:delta1}\\
\Delta_2
&=
\frac{2}{N(N-1)}
\sum_{i<j}
\frac{|\widehat c_{ij}-c_{ij}^{\mathrm{ref}}|}{|c_{ij}^{\mathrm{ref}}|}.
\label{eq:delta2}
\end{align}
Because the normalization is relative, small reference moments can amplify these metrics.  The exact-reference bands in the figures quantify the resulting finite-$S$ floor.

The figures also show two secondary diagnostics.  The negative pseudo-log-likelihood is
\begin{equation}
\mathrm{NPLL}
=-\frac1S\sum_{r=1}^S\sum_{i=1}^N
\log\target(\sigma_i^{(r)}\mid\sigma_{\setminus i}^{(r)}),
\label{eq:npll}
\end{equation}
where the Ising singleton conditional is analytic.  A plug-in KL proxy is based on
\begin{equation}
\KL(q\|\target)
=
\E_q[\log q(\sigma)+E(\sigma)]+\log Z.
\label{eq:klproxyidentity}
\end{equation}
The empirical plug-in entropy term depends on sample support and therefore on $S$; comparisons of this proxy are meaningful only at matched sample size.

\subsection{Results: conditional structure controls finite-budget generation}

\begin{figure}[t]
\centering
\includegraphics[width=\textwidth]{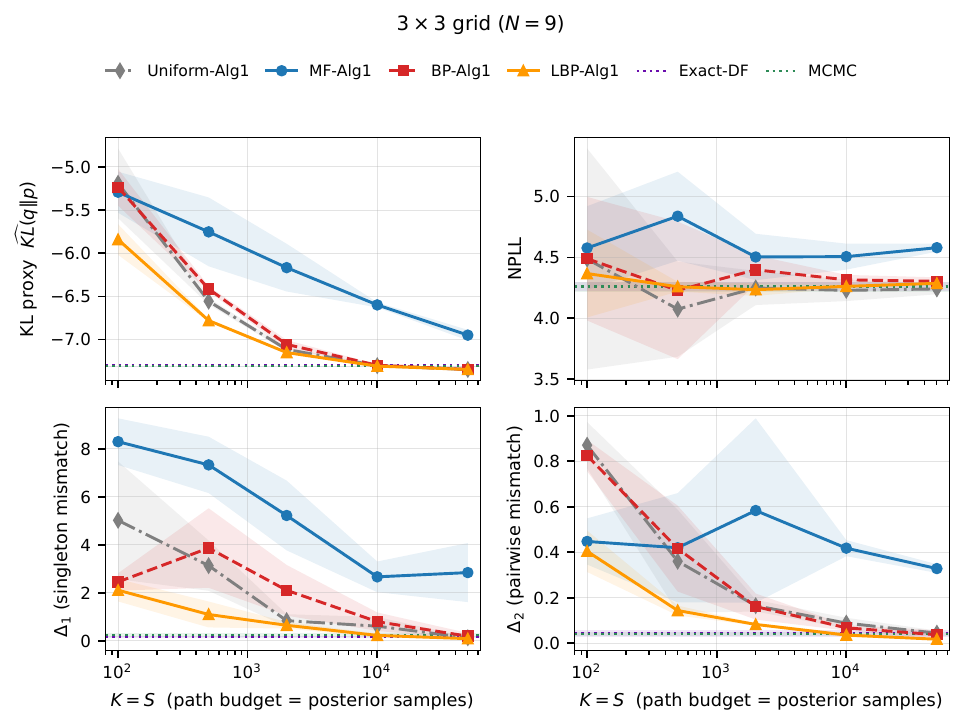}
\caption{$3\times3$ grid ($N=9$).  Route-resolved finite-particle Sampling Decisions sampler with $S=K$ for uniform, MF, BP, and LBP proposals; shading denotes one standard deviation over three seeds.  ``Exact-DF'' in the legend is the exact target i.i.d. reference at $S=5000$.  Uniform and BP approach the reference at the largest budget, MF remains limited by severe weight collapse, and LBP reaches the reference band at substantially smaller $K$.}
\label{fig:sweep3}
\end{figure}

Fig.~\ref{fig:sweep3} shows the $3\times3$ results.  Theorem~\ref{thm:iom} establishes that uniform, MF, and BP IOM priors have the same population limit.  Their visibly different finite-$K$ curves are therefore proposal effects.  Uniform and BP have nearly identical global ESS fractions and similar accuracy.  Naive MF is much worse: its fixed point is strongly polarized, its product law under-covers configurations important to the target, and the importance weights collapse.  Thus more accurate BP singleton marginals do not produce a substantially better proposal than uniform marks; the missing information is correlated structure.

Table~\ref{tab:ess} gives the global ESS fractions at $K=2\times10^4$.  LBP retains $0.884$, $0.772$, and $0.591$ of the nominal path budget on the three grids, respectively.  The product proposals range from $2.7\times10^{-4}$ to $9.2\times10^{-2}$.  The distinction is structural.  IOM proposals must represent the correlated target through an axis-aligned product terminal law, whereas LBP inserts revealed interactions during generation and removes the full energy factor from the path weight.

\begin{table}[t]
\centering
\caption{Global importance-weight effective sample size per path, $\ESS/K$, at $K=2\times10^4$.  The quantity broadly tracks finite-budget accuracy, although prefix-conditioned ESS is the sharper diagnostic.}
\label{tab:ess}
\begin{tabular}{lccc}
\toprule
Proposal & $3\times3$ ($N=9$) & $4\times4$ ($N=16$) & $5\times5$ ($N=25$)\\
\midrule
Uniform IOM & $0.086$ & $0.018$ & $0.008$\\
MF IOM      & $2.7\times10^{-4}$ & $5.4\times10^{-3}$ & $6.0\times10^{-4}$\\
BP IOM      & $0.092$ & $0.0045$ & $0.008$\\
LBP          & $\mathbf{0.884}$ & $\mathbf{0.772}$ & $\mathbf{0.591}$\\
\bottomrule
\end{tabular}
\end{table}

\begin{figure}[t]
\centering
\includegraphics[width=\textwidth]{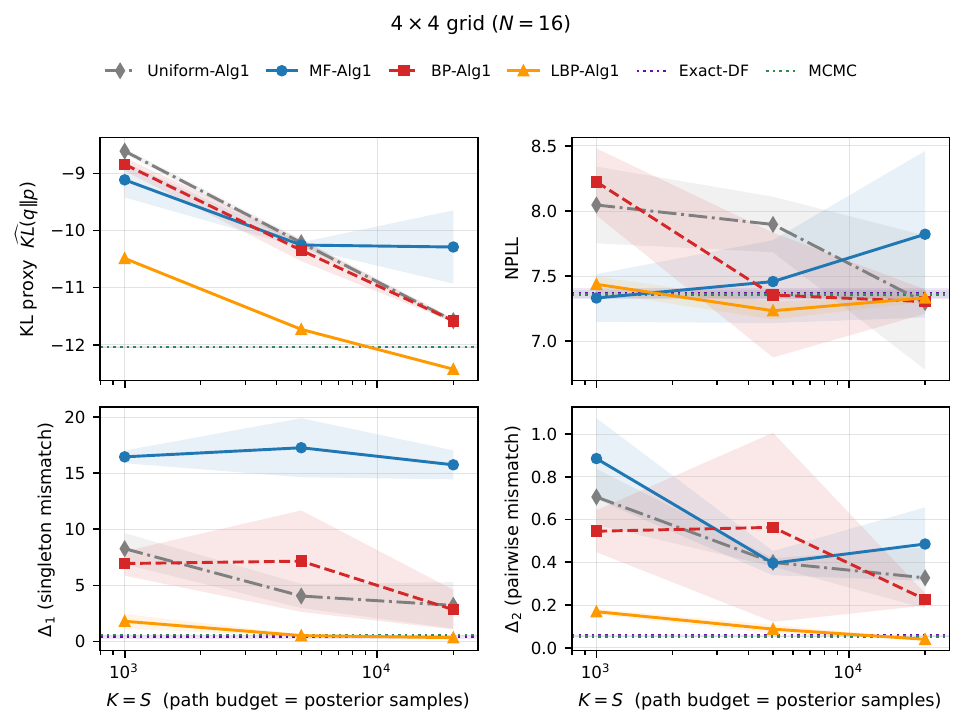}
\caption{$4\times4$ grid ($N=16$), with the same protocol as Fig.~\ref{fig:sweep3}.  LBP reaches the exact target reference band at $K=S=2\times10^4$.  Uniform and BP improve more slowly, while naive MF remains strongly affected by weight degeneracy.}
\label{fig:sweep4}
\end{figure}

\begin{figure}[t]
\centering
\includegraphics[width=\textwidth]{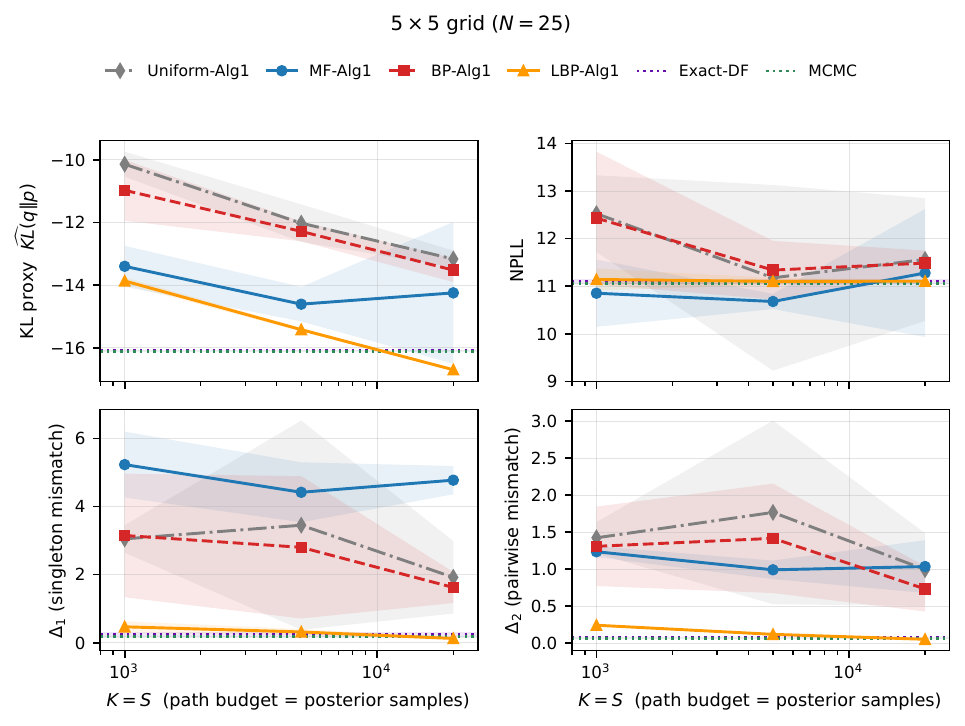}
\caption{$5\times5$ grid ($N=25$, $2^{25}\approx3.4\times10^7$ configurations), with exact references obtained by enumeration.  LBP lies on the exact target reference band on all four diagnostics at $K=S=2\times10^4$ in the reported runs.}
\label{fig:sweep5}
\end{figure}

Figs.~\ref{fig:sweep4} and \ref{fig:sweep5} show the larger grids.  At the largest tested budget, the LBP estimator is statistically consistent with the exact target reference on all four plotted diagnostics.  The product proposals improve much more slowly and remain outside the reference band in the moment metrics.  Across these three fixed instances, the LBP ESS fraction deteriorates much more mildly with $N$ than the product-prior ESS fractions.  This is promising evidence of variance reduction, not a proof of favorable asymptotic scaling.

\begin{figure}[t]
	\centering
	\includegraphics[width=\textwidth]{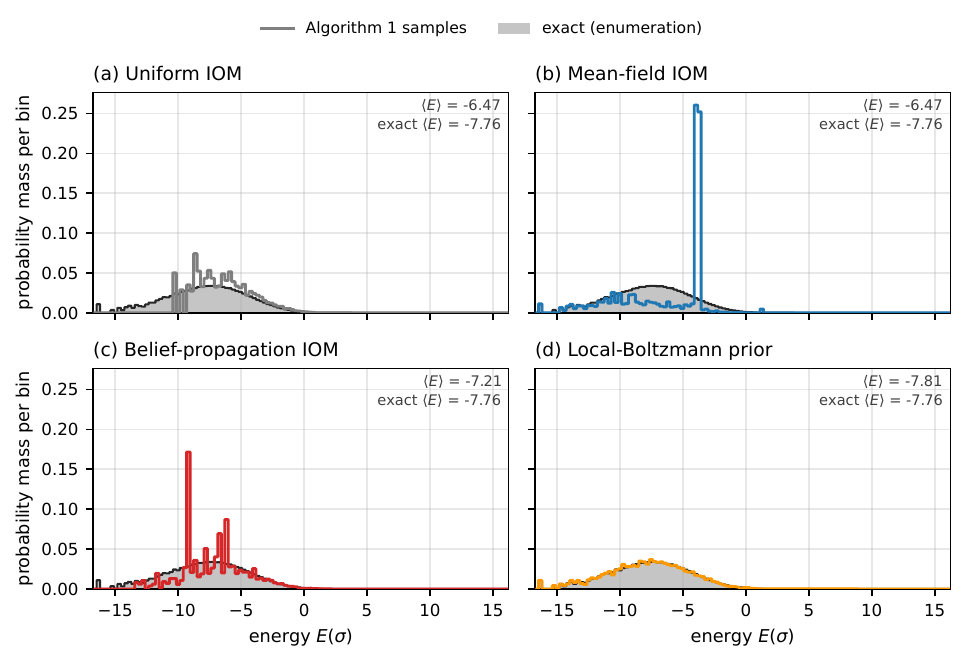}
	\caption{Energy-distribution comparison on the $5\times5$ instance
		at the largest common budget, $K=S=2\times10^{4}$.
		Each panel shows the binned energy probability-mass distribution of one
		finite-$K$ corrected sampler (colored) against the exact target
		distribution obtained by full enumeration of the $2^{25}$ states (gray);
		all panels share identical bins and axes, and report the sample mean
		energy versus the exact $\langle E\rangle=-7.76$. The panel ordering
		mirrors the ESS hierarchy of Table~\ref{tab:ess}: the LBP-Alg1 histogram
		(d) is indistinguishable from the exact law, uniform (a) is shifted
		toward higher energies, and the spikes in (b) and (c) are the fingerprint
		of importance-weight degeneracy --- a few dominant paths supply most of
		the posterior mass (ESS/$K\approx6\times10^{-4}$ and $8\times10^{-3}$,
		respectively). Agreement of this one-dimensional marginal is intuitive
		but not sufficient; it should be interpreted together with the singleton
		and pairwise discrepancies in
		Figures~\ref{fig:sweep3}--\ref{fig:sweep5}.}
	\label{fig:energy-distributions}
\end{figure}

Table~\ref{tab:delta} summarizes the singleton mismatch at the largest swept budget.  Values below the exact-target reference realization should not be interpreted as outperforming exact sampling: both candidate and reference metrics are computed from finite sample sets and fluctuate around their population values.

\begin{table}[t]
\centering
\small
\caption{Relative singleton mismatch $\Delta_1$ at the largest swept budget (mean $\pm$ standard deviation over three seeds).  The exact target and MCMC references use $S=5000$.}
\label{tab:delta}
\resizebox{\textwidth}{!}{%
\begin{tabular}{lccc}
\toprule
Method & $3\times3$, $K=S=5\times10^4$ & $4\times4$, $K=S=2\times10^4$ & $5\times5$, $K=S=2\times10^4$\\
\midrule
Uniform particle correction & $0.129\pm0.030$ & $3.22\pm2.07$ & $1.92\pm1.06$\\
MF particle correction      & $2.84\pm1.23$   & $15.74\pm1.29$ & $4.77\pm0.42$\\
BP particle correction      & $0.193\pm0.135$ & $2.84\pm1.77$ & $1.62\pm0.46$\\
LBP particle correction     & $\mathbf{0.091\pm0.020}$ & $\mathbf{0.341\pm0.091}$ & $\mathbf{0.123\pm0.032}$\\
\midrule
Exact target i.i.d. & $0.179\pm0.042$ & $0.382\pm0.200$ & $0.234\pm0.075$\\
MCMC                & $0.239\pm0.074$ & $0.581\pm0.027$ & $0.178\pm0.028$\\
\bottomrule
\end{tabular}}
\end{table}

For generative modeling, the controlled experiment isolates a design lesson that would be obscured in a large benchmark: matching easy low-order statistics is not the same as constructing a proposal with useful path-space overlap. The substantial gain from LBP comes from a physics-informed, prefix-dependent factorization that reduces the residual correction before any learning or resampling is introduced.

\subsection{Extension beyond exact enumeration.} 

The exactly enumerable grids isolate the mechanism under a gold-standard target reference. Appendix~\ref{app:10x10} asks whether the same proposal hierarchy persists when exact enumeration is no longer available. On a $10\times10$ Ising instance ($N=100$), LBP retains $\ESS/K=0.051$ at $K=5\times10^4$, whereas all three product proposals fall below $10^{-3}$. Already at $K=2\times10^3$, LBP achieves $\Delta_1=0.067$ and $\Delta_2=0.055$, outperforming every product proposal at $K=5\times10^4$. At the largest budget, its NPLL matches the long-run MCMC baseline to the reported precision, while its moment discrepancies remain larger: $(\Delta_1,\Delta_2)=(0.023,0.012)$ for LBP versus $(0.006,0.006)$ for MCMC. The larger experiment therefore extends the variance-reduction mechanism beyond the enumerated regime, but it does not establish a cost-normalized advantage over MCMC or favorable asymptotic scaling.

\section{Scope, limitations, and implications for learned generators}
\label{sec:limitations}

The present work establishes an exact architecture and a controlled test of its mechanism; it does not claim a state-of-the-art high-dimensional generator. Population exactness on a finite growing state graph should not be confused with computational tractability. The canonical correction can be as hard as evaluating target conditionals or prior terminal marginals, and the route-resolved estimator can still require a path budget that grows rapidly with dimension. Local-Boltzmann guidance reduces degeneracy on the tested ferromagnetic instances, but we make no polynomial-scaling claim. The exactly enumerable Ising grids are used because they allow the population law, the finite-particle approximation, and the diagnostics to be separated without an uncontrolled reference.

The analytical separation between prior policy and exact correction nevertheless defines a concrete route to learned scientific generators. In a larger system, the desirability $h_t$ or the look-ahead field $\widetilde h_i$ could be parameterized by a neural network and trained from weighted trajectories, Bellman residuals, or flow-consistency conditions. Prefix-conditioned resampling could be added when the particle bank degenerates, and the ordering policy could itself be learned. These extensions would amortize the exact path-space object identified here; they are future algorithms rather than results claimed in the present paper.

Sampling Decisions therefore complements, rather than replaces, resampling-based sequential Monte Carlo, learned GFlowNets, diffusion or bridge models, and MCMC. Its contribution is a common analytical reference that clarifies which part of a scientific generative model is supplied by physical structure, which part enforces global probabilistic consistency, and which part remains to be learned. Deployment in high-stakes settings would still require domain-specific validation, support and weight-collapse diagnostics, and explicit accounting of approximation error.

\section{Conclusion}
\label{sec:conclusion}

Sampling Decisions recasts scientific generation as two coupled modules: a prior policy that encodes inexpensive domain knowledge and a path-space correction that enforces the global target law. For Gibbs targets, the exact correction is the unique prior-relative KL projection and a Doob $h$-transform with a linear backward recursion. The same object is a desirability in linearly solvable control, a one-sided Schr\"odinger transport potential, and an ideal prefix value or flow for a compositional generator. The route-resolved formulation turns this population object into a direct, strongly consistent particle algorithm.

The cancellation theorem reveals a sharp representational principle. Within the IOM class, improving fixed singleton marginals cannot change the exact corrected dynamics; it can only change finite-budget weight variance. Useful guidance must encode conditional structure along the growing object. The Local-Boltzmann prior does so with a physics-informed factorization: revealed interactions are absorbed exactly, and the remaining path-space correction is the look-ahead free-energy field of the unrevealed subgraph. The controlled Ising experiments support this mechanism through substantially larger ESS and faster approach to the exact-target reference than product proposals. The $N=100$ extension shows that this hierarchy persists beyond exact enumeration: LBP remains operational and substantially closer to a long-run MCMC baseline, whereas product proposals undergo severe weight collapse. At the same time, the remaining gap in moment accuracy and the decrease of the LBP ESS fraction identify the nonlocal look-ahead information that must ultimately be learned or stabilized by resampling in a scalable implementation.

The broader implication for generative AI is not that one particular system particle sampler is already scalable. It is that the roles of physics, applied mathematics, and learning can be separated exactly. Statistical physics contributes energies, locality, and mechanistic inductive bias; stochastic control and transport contribute global probabilistic consistency; learning can then be reserved for amortizing the residual value or flow that is genuinely nonlocal. A sequential generator should therefore be judged not only by terminal samples or local marginals, but by how much target action it absorbs along each prefix and by the variance of the residual path correction. This provides a principled design criterion for future physics-informed GFlowNets, auto-regressive models, and discrete bridge samplers.

\begin{acknowledgments}
Automated language and coding assistants were used for editorial support and software organization. All mathematical derivations, numerical results, and scientific conclusions were checked by the authors, who take full responsibility for the content of the paper.

\end{acknowledgments}

\section*{Data and code availability}
The implementation and the data needed to reproduce the reported experiments are publicly available at \url{https://github.com/hamidrezabehjoo/SamplingDecisions}.

\appendix

\section{Additional proofs and identities}
\label{app:proofs}

\subsection{Terminal convergence from transition convergence}

For completeness, write the terminal law associated with a transition family $p$ as
\begin{equation}
P_T^p(\sigma)
=
\sum_{s_1,\ldots,s_{T-1}}
\prod_{t=0}^{T-1}p_t(s_{t+1}\mid s_t),
\qquad s_T=\sigma.
\label{eq:terminalpolynomial}
\end{equation}
On finite state spaces this is a multivariate polynomial in the transition entries.  Simultaneous convergence of all relevant entries therefore implies pointwise terminal convergence.  Because the terminal space is finite,
\[
\|P_T^{\widehat p_K}-P_T^p\|_{\TV}
=\frac12\sum_{\sigma\in\cX}
|P_T^{\widehat p_K}(\sigma)-P_T^p(\sigma)|\to0.
\]
This is the final step used in Theorem~\ref{thm:consistency}.

\subsection{Uniform-route chain rule}

Under the path target $P^{\mathrm{u}}(\pi,\sigma)=\target(\sigma)/N!$, the permutation is independent of $\sigma$.  Conditional on the revealed set $V_t$, every remaining node is equally likely to be next, giving probability $1/(N-t)$.  Conditional additionally on selecting node $i$, the spin value is distributed according to the ordinary target conditional given the revealed assignments.  This proves Eq.~\eqref{eq:uniformroutechainrule}.

\section{Canonical terminal-only correction for the Local-Boltzmann prior}
\label{app:canonicalLBP}

Eq.~\eqref{eq:lbppathlaw} implies
\begin{align}
\prior_{\mathrm{LBP},T}(\sigma)
&=
\sum_{\pi}\prior_{\mathrm{LBP}}(\pi,\sigma)
=
e^{-E(\sigma)}A_0(\sigma),
\label{eq:appA0}\\
A_0(\sigma)
&=
\frac1{N!}\sum_{\pi}
\frac1{Z_\pi^{\mathrm{loc}}(\sigma)}.
\label{eq:appA0def}
\end{align}
Since the terminal marginal is normalized, $\sum_\sigma e^{-E(\sigma)}A_0(\sigma)=1$.  The canonical terminal ratio from Eq.~\eqref{eq:terminalweight} is
\begin{equation}
w_T^{\mathrm{can}}(\sigma)
=
\frac{e^{-E(\sigma)}}{\prior_{\mathrm{LBP},T}(\sigma)}
=
\frac1{A_0(\sigma)}.
\label{eq:canonicalLBPweight}
\end{equation}
The prior conditional distribution over orderings is
\begin{equation}
\prior_{\mathrm{LBP}}(\pi\mid\sigma)
=
\frac{1}{N!\,Z_\pi^{\mathrm{loc}}(\sigma)A_0(\sigma)}.
\label{eq:canonicalorderconditional}
\end{equation}
The canonical correction therefore has joint path law
\begin{equation}
P^{\star}_{\mathrm{LBP}}(\pi,\sigma)
=
\target(\sigma)
\frac{1}{N!\,Z_\pi^{\mathrm{loc}}(\sigma)A_0(\sigma)}.
\label{eq:canonicalLBPpath}
\end{equation}
Unlike the uniform-route target, Eq.~\eqref{eq:canonicalLBPpath} generally correlates ordering and terminal configuration.  The canonical process may therefore reweight both node selection and spin assignment.  It is the unique KL-minimizer relative to LBP under the terminal constraint alone, but direct evaluation requires the order average $A_0(\sigma)$.  The route-resolved implementation in the main text trades that terminal-only optimality for an analytically specified uniform route law and tractable pathwise weights.

\section{Mean-field and belief-propagation singleton priors}
\label{app:mf-bp}

The IOM experiments use uniform node order and differ only in the fixed singleton marks $q_i$.  For a magnetization estimate $m_i\in(-1,1)$, the corresponding mark is
\begin{equation}
q_i(a)=\frac{1+a m_i}{2},
\qquad a\in\{-1,+1\}.
\label{eq:markfromm}
\end{equation}

\paragraph{Naive mean field.}
The MF magnetizations solve the fixed-point equations
\begin{equation}
m_i
=
\tanh\!\left(h_i+\sum_{j:(i,j)\in\mathcal E}J_{ij}m_j\right).
\label{eq:mffixedpoint}
\end{equation}
The implementation iterates these equations to a fixed point and uses Eq.~\eqref{eq:markfromm}.  In the reported ferromagnetic instances, the selected fixed point can be highly polarized, which makes the product proposal overconfident.

\paragraph{Belief propagation.}
For pairwise Ising models, write cavity fields $H_{j\to i}$ satisfying
\begin{equation}
H_{j\to i}
=
h_j+
\sum_{k\in\partial j\setminus i}
\operatorname{atanh}\!\left(
\tanh(J_{jk})\tanh(H_{k\to j})
\right).
\label{eq:bpfields}
\end{equation}
The BP singleton estimate is
\begin{equation}
m_i^{\mathrm{BP}}
=
\tanh\!\left(
h_i+
\sum_{j\in\partial i}
\operatorname{atanh}\!\left(
\tanh(J_{ij})\tanh(H_{j\to i})
\right)
\right),
\label{eq:bpsingleton}
\end{equation}
followed by Eq.~\eqref{eq:markfromm}.  On loopy graphs these are approximate marginals, but Theorem~\ref{thm:iom} does not depend on their accuracy: all fixed marks cancel from the population corrected dynamics.

\clearpage
\section{$10 \times 10$}

\pgfplotsset{compat=1.18}

\begin{table}[t]
\centering
\caption{%
$10\times 10$ Ising model ($N=100$, sub-critical couplings $\beta J=0.28$).
Comparison of route-resolved finite-particle Sampling Decisions
(Algorithm~1) with uniform, mean-field (MF), belief-propagation (BP),
and Local-Boltzmann (LBP) priors.
The MCMC reference uses $5\times 10^{4}$ samples with burn-in $2\times 10^{4}$
and thinning $50$.
ESS fractions below $10^{-3}$ are reported as $<$$10^{-3}$.
}
\label{tab:10x10}
\small
\begin{tabular}{
  @{} l 
  c
  c
  S[table-format=2.1]
  S[table-format=1.3]
  S[table-format=1.3]
  S[table-format=2.1]
  @{}
}
\toprule
{Prior} & {$K$} & {ESS/$K$} & {KL} & {$\Delta_1$} & {$\Delta_2$} & {NPLL} \\
\midrule
Uniform & 2000  & 0.002      & -18.5 & 0.429 & 0.434 & 65.2 \\
        & 5000  & $<$$10^{-3}$ & -21.5 & 0.737 & 0.610 & 67.1 \\
        & 20000 & $<$$10^{-3}$ & -24.4 & 0.795 & 0.618 & 56.2 \\
        & 50000 & $<$$10^{-3}$ & -23.6 & 0.196 & 0.267 & 60.3 \\
\midrule
MF      & 2000  & 0.001      & -23.8 & 0.944 & 0.825 & 73.8 \\
        & 5000  & $<$$10^{-3}$ & -23.9 & 0.835 & 0.700 & 76.4 \\
        & 20000 & $<$$10^{-3}$ & -25.6 & 0.673 & 0.543 & 69.5 \\
        & 50000 & $<$$10^{-3}$ & -21.8 & 0.852 & 0.761 & 72.6 \\
\midrule
BP      & 2000  & 0.002      & -30.5 & 0.509 & 0.369 & 52.1 \\
        & 5000  & 0.004      & -31.4 & 0.369 & 0.197 & 54.7 \\
        & 20000 & $<$$10^{-3}$ & -31.9 & 0.437 & 0.339 & 49.7 \\
        & 50000 & $<$$10^{-3}$ & -34.7 & 0.357 & 0.222 & 48.8 \\
\midrule
LBP     & 2000  & 0.083      & -42.3 & 0.067 & 0.055 & 40.9 \\
        & 5000  & 0.093      & -42.8 & 0.035 & 0.035 & 41.5 \\
        & 20000 & 0.050      & -44.7 & 0.035 & 0.020 & 40.9 \\
        & 50000 & 0.051      & -45.9 & 0.023 & 0.012 & 40.6 \\
\midrule
MCMC (ref) & 50000 & {---} & -47.8 & 0.006 & 0.006 & 40.6 \\
\bottomrule
\end{tabular}
\end{table}


\section{Extension beyond exact enumeration: a $10\times10$ Ising instance}
\label{app:10x10}

At scales where exact enumeration is unavailable, Algorithm~\ref{alg:particleDF} remains valid with a local-Gibbs starvation fallback. Table~\ref{tab:10x10} reports a $10\times 10$ Ising instance ($N=100$, sub-critical couplings $\langle J\rangle=0.28$) against a long-MCMC reference ($5\times 10^{4}$ samples, burn-in $2\times 10^{4}$, thin $50$).

LBP achieves a KL proxy of $-45.9$ with the NPLL matching the MCMC baseline to the reported precision and the KL proxy lying closer to the MCMC value than those of the product proposals ($-47.8$ and $40.6$ respectively), and its absolute moment errors converge to within $4\times$ of the reference at $K=5\times 10^{4}$. The product priors remain one to two orders of magnitude worse in moment space, with effective sample size collapsing below $10^{-3}$ and KL values $15$--$25$ units away from the reference.

\begin{figure*}[t]
\centering
\begin{tikzpicture}
\begin{groupplot}[
  group style={
    group size=2 by 2,
    xlabels at=edge bottom,
    ylabels at=edge left,
    horizontal sep=2.2em,
    vertical sep=2em,
  },
  width=0.45\textwidth,
  height=0.38\textwidth,
  grid=major,
  grid style={line width=0.2pt, draw=gray!30},
  tick label style={font=\small},
  label style={font=\small},
]

\nextgroupplot[
  ylabel={ESS / $K$},
  xmode=log,
  ymode=log,
  xmin=1500, xmax=70000,
  ymin=1e-4, ymax=1,
]
\addplot[color=gray, mark=diamond*, mark options={scale=1.2}, thick]
  coordinates {(2000,0.002) (5000,0.0005) (20000,0.0003) (50000,0.0002)};
\addplot[color=blue, mark=*, thick]
  coordinates {(2000,0.001) (5000,0.0004) (20000,0.0002) (50000,0.0002)};
\addplot[color=red, mark=square*, thick]
  coordinates {(2000,0.002) (5000,0.004) (20000,0.0003) (50000,0.0002)};
\addplot[color=orange, mark=triangle*, mark options={scale=1.3}, very thick]
  coordinates {(2000,0.083) (5000,0.093) (20000,0.050) (50000,0.051)};

\nextgroupplot[
  ylabel={KL proxy $\widehat{\mathrm{KL}}(q\|\mu)$},
  ylabel style={at={(rel axis cs:1.22,0.25)}, anchor=west, rotate=-0},
    yticklabel pos=right,
  xmode=log,
  xmin=1500, xmax=70000,
  ymin=-50, ymax=-15,
  legend to name=leg:10x10,
  legend style={legend columns=5, draw=none, fill=none, font=\footnotesize, /tikz/every even column/.append style={column sep=0.5em}},
]
\addplot[color=gray, mark=diamond*, mark options={scale=1.2}, thick]
  coordinates {(2000,-18.5) (5000,-21.5) (20000,-24.4) (50000,-23.6)};
\addplot[color=blue, mark=*, thick]
  coordinates {(2000,-23.8) (5000,-23.9) (20000,-25.6) (50000,-21.8)};
\addplot[color=red, mark=square*, thick]
  coordinates {(2000,-30.5) (5000,-31.4) (20000,-31.9) (50000,-34.7)};
\addplot[color=orange, mark=triangle*, mark options={scale=1.3}, very thick]
  coordinates {(2000,-42.3) (5000,-42.8) (20000,-44.7) (50000,-45.9)};
\addplot[color=black, dashed, line width=1.5pt]
  coordinates {(1500,-47.8) (70000,-47.8)};
\legend{Uniform, MF, BP, LBP, MCMC }

\nextgroupplot[
  xlabel={Path budget $K$},
  ylabel={$|\Delta_{1}|$},
  xmode=log,
  xmin=1500, xmax=70000,
  ymin=0, ymax=1.05,
]
\addplot[color=gray, mark=diamond*, mark options={scale=1.2}, thick]
  coordinates {(2000,0.429) (5000,0.737) (20000,0.795) (50000,0.196)};
\addplot[color=blue, mark=*, thick]
  coordinates {(2000,0.944) (5000,0.835) (20000,0.673) (50000,0.852)};
\addplot[color=red, mark=square*, thick]
  coordinates {(2000,0.509) (5000,0.369) (20000,0.437) (50000,0.357)};
\addplot[color=orange, mark=triangle*, mark options={scale=1.3}, very thick]
  coordinates {(2000,0.067) (5000,0.035) (20000,0.035) (50000,0.023)};
\addplot[color=black, dashed, line width=1.5pt]
  coordinates {(1500,0.0058) (70000,0.0058)};

\nextgroupplot[
  xlabel={Path budget $K$},
  ylabel={$|\Delta_2|$},
  ylabel style={at={(rel axis cs:1.22,0.35)}, anchor=west, rotate=-0},
  yticklabel pos=right,
  xmode=log,
  xmin=1500, xmax=70000,
  ymin=0, ymax=0.9,
]
\addplot[color=gray, mark=diamond*, mark options={scale=1.2}, thick]
  coordinates {(2000,0.434) (5000,0.610) (20000,0.618) (50000,0.267)};
\addplot[color=blue, mark=*, thick]
  coordinates {(2000,0.825) (5000,0.700) (20000,0.543) (50000,0.761)};
\addplot[color=red, mark=square*, thick]
  coordinates {(2000,0.369) (5000,0.197) (20000,0.339) (50000,0.222)};
\addplot[color=orange, mark=triangle*, mark options={scale=1.3}, very thick]
  coordinates {(2000,0.055) (5000,0.035) (20000,0.020) (50000,0.012)};
\addplot[color=black, dashed, line width=1.5pt]
  coordinates {(1500,0.0063) (70000,0.0063)};

\end{groupplot}

\node[anchor=south, yshift=1.2em] at ($(group c1r1.north west)!0.5!(group c2r1.north east)$) {\ref{leg:10x10}};

\end{tikzpicture}
\caption{%
$10\times 10$ Ising model ($N=100$, sub-critical couplings).
The dashed line in \textbf{(b--d)} marks the long-MCMC reference. LBP moves systematically toward the long-run MCMC baseline over the tested budgets, whereas the product proposals remain strongly weight-degenerate and substantially less accurate.}
\label{fig:10x10}
\end{figure*}

The exactly enumerable grids are used to separate the population law, the finite-particle approximation, and diagnostic noise under a gold-standard reference. Appendix~\ref{app:10x10} extends the experiment to one $N=100$ instance, where exact enumeration is unavailable and a long-run MCMC calculation is used as an operational baseline. Because the coupling regime also differs, this extension is not a controlled scaling study. It shows that the proposal hierarchy persists beyond enumeration, while the remaining ESS decay precludes any polynomial-scaling claim.

\clearpage                                            
\bibliography{bib/MishaPapers, bib/StochasticOptimalControl, bib/DiffusionModels,bib/combinatorics}
\bibliographystyle{unsrt}

\end{document}